\ifdefined\XeTeXrevision
\else
\pdfoutput=1
\fi
\documentclass{lynnreal}

\usepackage{amsmath,amssymb}   
\usepackage{pifont}            

\microtypesetup{expansion=false}

\usepackage{xspace}
\makeatletter
\DeclareRobustCommand\onedot{\futurelet\@let@token\@onedot}
\def\@onedot{\ifx\@let@token.\else.\null\fi\xspace}

\makeatother

\newcommand{\vx}{\mathbf{x}}
\newcommand{\vc}{\mathbf{c}}
\newcommand{\va}{\mathbf{a}}
\newcommand{\KL}{\mathrm{KL}}

\newcommand{\vepsilon}{\boldsymbol{\epsilon}}

\newcommand{\ours}{\textcolor{lynnaccent}{\textbf{BiWM}}}

\let\cite\citep

\title{{\LARGE BiWM: Advancing Open-Source Interactive Video\\ World Models with Bidirectional Autoregression}}

\author[1,2,3*\ddagger]{Shaohao Rui}
\author[2,4*]{Xiaofeng Mao}
\author[1]{Zhanyu Zhang}
\author[1,2]{Peijia Lin}
\author[1]{Yansong Zhu}
\author[1,2]{Yibo Zhang}
\author[1]{Haibin Wan}
\author[1]{Zhangrui Zhao}
\author[1,2,4\dagger]{Weijie Ma}

\affiliation[1]{LynnReal AI}
\affiliation[2]{Shanghai Innovation Institute}
\affiliation[3]{Shanghai Jiao Tong University}
\affiliation[4]{Fudan University}

\contribution[*]{Equal contribution.}
\contribution[\ddagger]{Project Lead.}
\contribution[\dagger]{Corresponding Author.}

\abstract{
Interactive video world models commonly convert bidirectional video generators into causal autoregressive
systems through control fine-tuning, autoregressive training, causal initialization, and few-step distillation.
This pipeline is costly, while frozen causal histories accumulate errors that degrade long-horizon fidelity and
controllability. We present \ours{}, the first open-source full-stack training framework for bidirectional
autoregressive video world models. \ours{} retains full attention within each generated chunk and requires only
two stages: camera/action-control fine-tuning and few-step Distribution Matching Distillation (DMD). Both stages
converge within a few hundred optimizer steps on $8\times$H200 GPUs. The framework supports
Wan2.1-T2V-1.3B, Wan2.2-TI2V-5B, HunyuanVideo-1.5-TI2V-8B, and LTX-2.3-22B, together with real-world camera
control. Supervised and forward-KL anchors
mitigate DMD mode collapse and preserve scene dynamics. \ours{} provides a compact, reproducible path from
pretrained bidirectional video models to interactive, controllable, and efficient world models.
}

\date{2026.06.09}
\correspondence{Weijie Ma at \email{weijiema@lynnreal.com}}

\lynndata[Code]{\url{https://github.com/LynnReal-AI/BiWM}}

\begin{document}

\maketitle

\begin{figure}[!h]
    \centering
    \vspace{-1mm}
    \includegraphics[width=\textwidth]{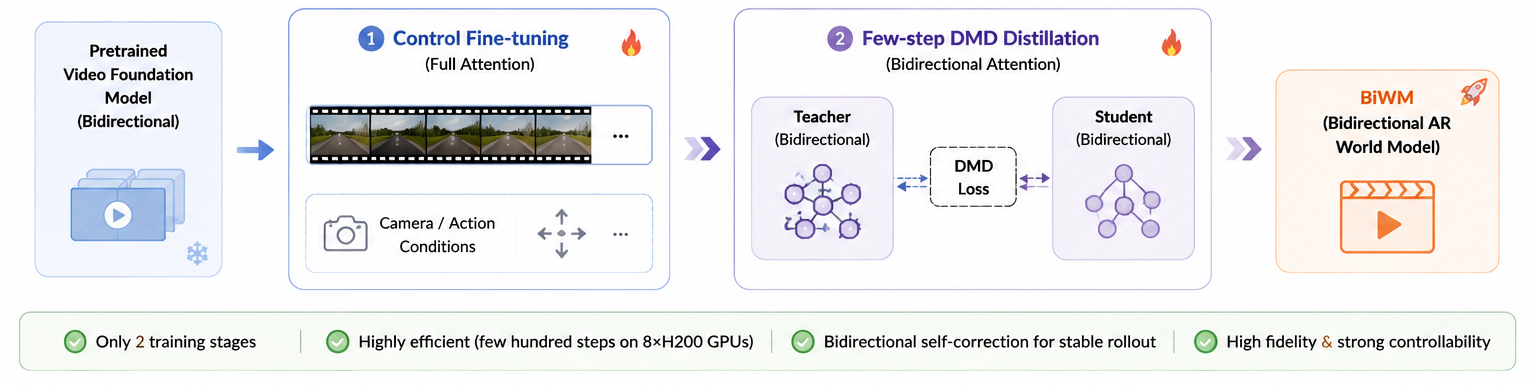}
    \vspace{-1mm}
    \caption{\textbf{Overview of \ours{}.} From a pretrained bidirectional video foundation model, \ours{}
    runs just two short training stages---camera/action control fine-tuning and few-step DMD distillation, both
    keeping full bidirectional attention---to obtain a bidirectional autoregressive interactive world model. The
    recipe uses only two training stages, is highly efficient (a few hundred steps on $8\times$H200 GPUs),
    self-corrects through bidirectional rollout for stable long-horizon generation, and attains high fidelity with
    strong controllability.}
    \label{fig:teaser}\label{fig:pipeline}\label{fig:architecture}
    \vspace{-2mm}
\end{figure}

\begin{figure*}[!t]
    \centering
    \includegraphics[width=\textwidth]{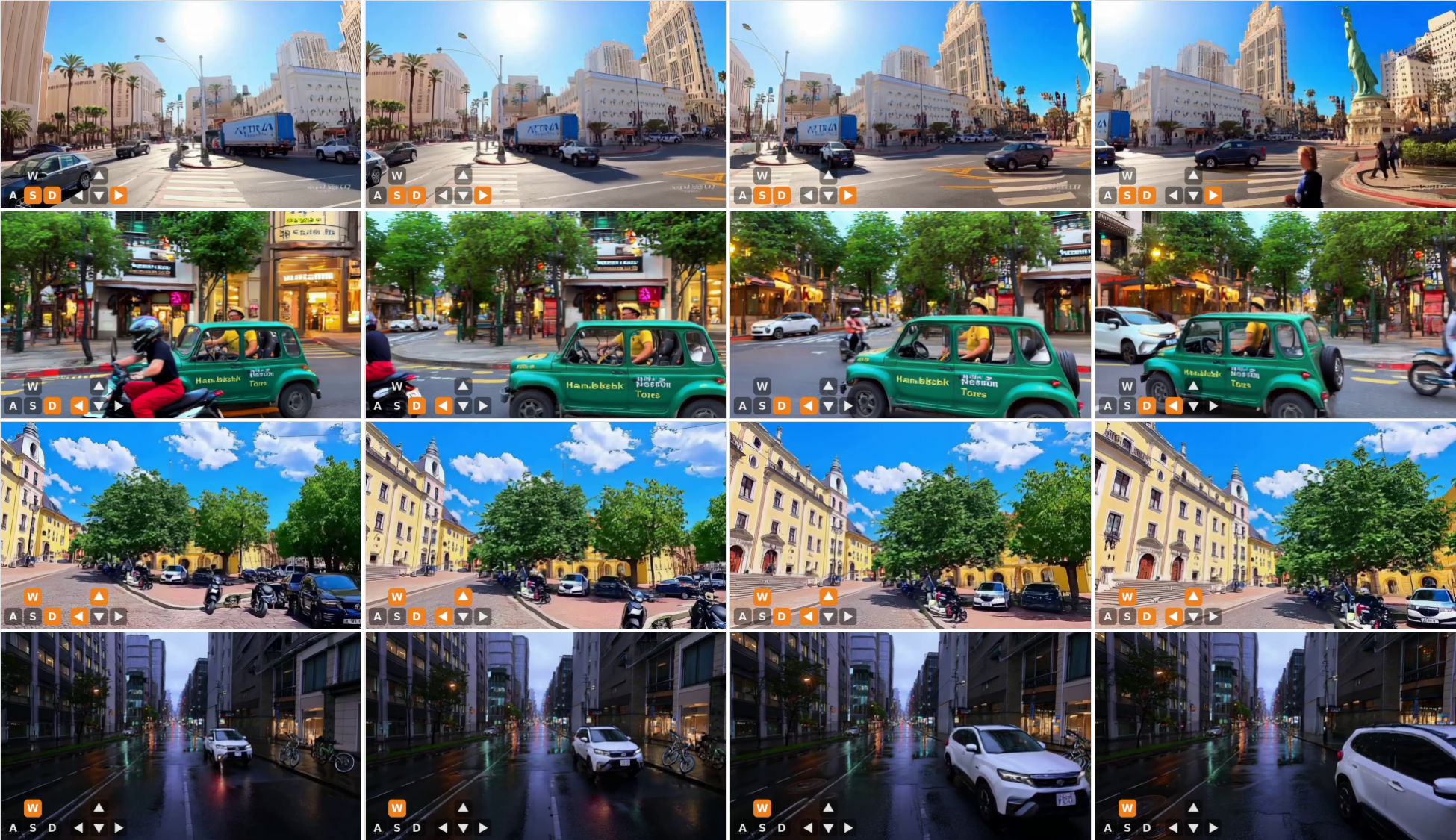}
    \vspace{-6pt}
    \caption{\textbf{Interactive world exploration with \ours{}.} Driven by discrete keyboard$+$mouse actions,
    \ours{} lets a user explore a generated world. Text-to-video rollouts on Sekai-domain street scenes, each row
    navigated under a different constant discrete camera action---from top: \textit{backward-right $+$ yaw-right},
    \textit{right $+$ yaw-left}, \textit{forward-right $+$ pitch-up, yaw-left}, and \textit{forward, static look}.
    The bottom-left joystick overlay shows the action; the camera obeys each prescribed translation and look
    direction while preserving scene fidelity.}
    \label{fig:camera_control}
\end{figure*}

\section{Introduction}

Given a text or image prompt, contemporary video-diffusion models synthesize seconds to minutes of
high-fidelity, temporally coherent footage~\citep{brooks2024video,bao2024vidu,yang2025cogvideox,wan2025wan,kong2024hunyuanvideo,hacohen2024ltx}.
Re-purposing such a generator into an interactive video world model, one that continues a virtual world
frame by frame under a stream of user actions (above all camera movements) and lets a person steer through it, has
become a central goal of generative world modeling~\citep{genie3,bruce2024genie,sun2025worldplay,tang2025hunyuan,mao2025yume,ye2025yan,he2025matrix,xiang2025pan}.
Turning such an offline bidirectional generator into an autoregressive one that emits frames on
demand is what makes these systems interactive and real-time, and is therefore the central technical
challenge in building them.

\begin{figure}[!t]
    \centering
    \includegraphics[width=\textwidth]{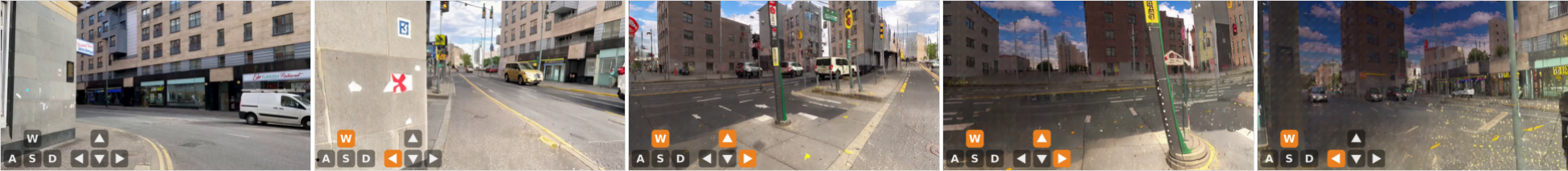}
    \vspace{-6pt}
    \caption{\textbf{Why fully causal camera control collapses.} A causal autoregressive baseline
    (Self-Forcing-style, continuous-pose control) rolling out an image-to-video clip under a walking camera
    trajectory (joystick overlay, bottom-left shows the commanded action). Left to right, errors frozen into the
    KV cache and drift in the camera response compound, and the scene degrades from a clean street into a washed-out,
    structureless frame. \ours{}'s chunk-wise bidirectional rollout (self-correcting history) and discrete text-camera
    control are designed to avoid both failure modes.}
    \label{fig:causal}
\end{figure}

Existing autoregressive world models fall into two families, distinguished by how the frames within a window
attend to one another. \textbf{Causal} models~\citep{team2026advancing,hong2025relic,nam2026worldcam,hunyuanworld2025hy}
impose a causal mask so that each frame sees only its past;\footnote{In practice many ``causal'' world models adopt
a local window: attention is bidirectional within a short window of frames and causal only
across windows, a compromise that recovers some visual quality. For clarity of exposition we set this
refinement aside and speak of fully causal rollout; it does not affect the argument, since a window's
representation is still frozen once it leaves the cache.} their appeal is efficiency, since the past can be
stored as a key--value (KV) cache and reused to accelerate the rollout. This caching, however, conceals a
structural weakness: once a frame is frozen into the KV cache its representation can never be revised, so any error
in the generated history is permanent and compounds as the rollout lengthens, eventually corrupting both the
imagery and the model's response to control. The drift is more damaging for video diffusion than for language. An
autoregressive language model predicts over a discrete vocabulary and re-quantizes onto valid tokens at
every step, which endows it with an innate ability to absorb and correct small mistakes; a diffusion-based video
model instead fits a continuous distribution over pixels, where sub-token deviations are never snapped back and
accumulate unchecked until the scene collapses. The two failures compound: errors in the imagery and drift in the
camera response reinforce one another, so a causal rollout under camera control degrades faster than either alone
(Fig.~\ref{fig:causal}). \textbf{Bidirectional} models, namely the
Yume series~\citep{mao2026yume1,mao2025yume} and Matrix-Game-3.0~\citep{wang2026matrix}, instead let every frame in
a window attend to every other, exactly as the pretrained backbone does, and this is precisely what counteracts the
drift: because earlier history latents remain visible to, and are refreshed alongside, the frames currently being
denoised, the model continually self-corrects its own past, trading a modest amount of caching efficiency for
substantially better fidelity and controllability over long horizons. What makes this trade-off practical is
few-step distillation: once a window denoises in only a handful of steps, retaining full bidirectional attention
within it incurs little additional cost, and error resilience, rather than latency, becomes the deciding factor.

These bidirectional systems confirm the benefit empirically, reporting sharper frames and more stable
long-horizon exploration than their causal counterparts. What the community still lacks, however, is an open,
end-to-end recipe for the paradigm. minWM~\citep{zhao2026minwm} has open-sourced a full-stack framework for
causal interactive world models, yet no full-stack, open-source counterpart exists for the
bidirectional autoregressive paradigm, which leaves its strong empirical results difficult to reproduce or extend.

We close this gap with \ours{}, to our knowledge the first full-stack, open-source framework for building
interactive video world models under the bidirectional autoregressive paradigm, designed to balance generation
quality against generation speed. \ours{} keeps the backbone's native full attention within each short
window of latent frames and pays the autoregressive cost only across windows, conditioning each window on
the history of those before it. Starting from a pretrained video foundation model, the recipe needs only two
stages: a first stage that injects camera control by fine-tuning, and a second stage that directly performs
few-step self-rollout DMD distillation, building on Self-Forcing~\citep{huang2026self} but in the
chunk-wise bidirectional rather than causal setting~\citep{yin2024one,yin2024improved}, after which the backbone becomes a camera- and
action-controllable interactive world model. To counter the mode-seeking tendency of distribution-matching
distillation, which otherwise collapses scene dynamics, we augment the DMD objective with anti-degradation
terms, including supervised and mass-covering forward-KL anchors that preserve detail and motion diversity.

Importantly, \ours{} is built for the resource budgets of academic research. It uses only two training stages,
compared with four in minWM, and is inexpensive to train: camera control and DMD distillation jointly converge
within a few hundred optimizer steps on $8\times$H200 GPUs, so that a complete world model can be validated and
iterated within hours rather than weeks. Because none of the recipe is tied to a particular backbone, we provide
full-stack training across architectures and modalities, including Wan2.1-T2V-1.3B,
Wan2.2-TI2V-5B~\citep{wan2025wan}, HunyuanVideo-1.5-TI2V-8B~\citep{kong2024hunyuanvideo,esser2024scaling},
and LTX-2.3-22B~\citep{hacohen2024ltx}. The same framework also supports secondary fine-tuning of
existing bidirectional autoregressive models such as Yume-1.5 and Matrix-Game-3.0, adapting them to new data
distributions at low cost, and it enables real-world camera control, a regime that proves nearly
uncontrollable under minWM. Fig.~\ref{fig:pipeline} summarizes the two-stage recipe. We position \ours{} not as a
competitor to causal frameworks but as their bidirectional complement in the same open-source design space,
trading a small amount of per-window latency for fidelity, controllability, and substantially shorter training.

\paragraph{\normalfont\bfseries Contributions.}
\begin{itemize}
    \renewcommand\labelitemi{$\bullet$}
    \item We introduce \ours{}, the first full-stack, open-source framework for interactive video world models
    under the \textbf{bidirectional autoregressive} paradigm, with full bidirectional attention within a chunk
    and autoregression across chunks, positioned as the bidirectional complement to causal frameworks such as minWM.

    \item We design a compact two-stage recipe (camera-control fine-tuning followed by few-step DMD
    distillation, versus four stages in minWM) that is deliberately suited to academic budgets: both stages jointly
    converge within a few hundred optimizer steps on $8\times$H200 GPUs. To prevent the mode-seeking collapse of
    distribution-matching distillation, we add supervised and mass-covering forward-KL anchors that preserve
    visual detail and scene dynamics.

    \item We demonstrate generality and reproducibility: \ours{} provides full-stack training across
    Wan2.1-T2V-1.3B, Wan2.2-TI2V-5B, HunyuanVideo-1.5-TI2V-8B, and LTX-2.3-22B, additionally supports secondary
    fine-tuning of bidirectional autoregressive models (Yume-1.5, Matrix-Game-3.0) to new data distributions, and
    enables real-world camera control that proves nearly uncontrollable under minWM. We release code, scripts, and
    checkpoints together with reproducible component studies.
\end{itemize}


\section{Related Work}

\paragraph{\normalfont\bfseries Video (and audio--video) diffusion backbones.}
Large-scale diffusion transformers have become the generative prior behind most recent video world models,
producing high-fidelity, temporally coherent clips across three broad architectural families: cross-attention
conditioned designs~\citep{wan2025wan,yang2025cogvideox,bao2024vidu}, MMDiT designs that jointly attend over text
and video tokens~\citep{esser2024scaling,kong2024hunyuanvideo}, and, increasingly, models that generate
synchronized audio and video~\citep{hacohen2024ltx}. A property they all share is full bidirectional
spatiotemporal attention over the entire clip---the very source of their fidelity. \ours{} takes such a model as
its teacher and retains this bidirectional attention within each generated chunk, rather than re-training
it to be strictly causal. Since our recipe alters only the conditioning and the rollout, leaving the backbone's
attention untouched, the same framework transfers cleanly across all three families.

\paragraph{\normalfont\bfseries Causal interactive world models.}
Most real-time world models convert an offline generator into a controllable, causal, low-latency
roll-out engine~\citep{genie3,bruce2024genie,sun2025worldplay,tang2025hunyuan,mao2025yume,ye2025yan,xiang2025pan,he2025matrix,hong2025relic,shin2025motionstream,feng2025vidarc},
typically following the block-causal AR template of CausVid~\citep{yin2025slow} and
Self-Forcing~\citep{huang2026self} and distilling it to a few steps. minWM~\citep{zhao2026minwm} packages this
conversion via Causal Forcing~\citep{zhu2026causal,zhao2026causal} with continuous PRoPE~\citep{li2026cameras}
camera control. \ours{} differs at the level of paradigm, keeping windows bidirectional, and in its controls
(discrete text-camera actions), its objective (multi-objective short distillation), and its breadth of
backbones.

\paragraph{\normalfont\bfseries Bidirectional interactive world models.}
A complementary line keeps each window bidirectional. The Yume series (Yume-1.0~\citep{mao2025yume} and
Yume-1.5~\citep{mao2026yume1}) and Matrix-Game-3.0~\citep{wang2026matrix} let every frame in a window attend to
every other and report sharper, more stable long-horizon rollouts than causal models, confirming the paradigm's
benefit empirically. These systems remain hard to reproduce, however: at the time of writing none has released its
training dataset or a complete training pipeline, and Matrix-Game-3.0 in particular still depends on a many-stage
training recipe. \ours{} targets exactly this gap, offering a fully open, two-stage recipe for the
bidirectional paradigm together with its data and training code.

\paragraph{\normalfont\bfseries Camera-signal injection.}
Existing ways to inject camera control into a video diffusion backbone fall into two families.
Absolute injection adds the camera signal onto the per-frame hidden state, either as a global,
low-frequency control applied to the latent before the DiT~\citep{he2024cameractrl}, or layer-by-layer into
every block's hidden state~\citep{team2026advancing}. Because the pose is encoded as an absolute per-frame signal,
this couples the temporal dynamics of the camera trajectory with those of the video itself, which
tends to amplify error accumulation over long rollouts. Relative injection instead alters the
inter-frame attention so that the interaction between two frames accounts for their relative pose;
representative methods include CaPE~\citep{kong2024eschernet}, GTA~\citep{miyato2024gta}, and
PRoPE~\citep{li2026cameras} (also adopted in HunyuanWorld~1.5~\citep{hunyuanworld2025hy}), as well as
UCPE~\citep{zhang2026unified} (adopted in SANA-WM). These are more robust, but they still inject the camera signal as a
residual branch alongside the original attention; even with zero initialization, the residual perturbs
the pretrained attention and induces a transient drop in visual quality early in training. Both families
typically require a relatively heavy camera encoder or per-layer learnable injection modules, converge slowly,
and lean on large batch sizes for training stability. In contrast, \ours{} casts camera control as a
conditioning task and injects the signal through the text space directly into the video tokens
(Sec.~\ref{sec:camtext}): it adds no new learnable parameters, converges within roughly a hundred steps,
and leaves the base generator's visual quality intact.

\paragraph{\normalfont\bfseries Few-step distillation and long-horizon memory.}
Distribution matching distillation (DMD)~\citep{wang2023prolificdreamer,luo2023diff,yin2024one,yin2024improved}
and consistency distillation~\citep{song2023consistency} compress many-step samplers to a handful of steps;
applied to AR video, DMD with self-rollout~\citep{yin2025slow,huang2026self,yin2025slow} is the standard route
to real-time generation. As its reverse-KL objective is mode-seeking, we pair it with a supervised regression
(SFT) term and mass-covering forward-KL anchors, which together stabilize convergence and preserve visual detail
and motion diversity.
For long-horizon rollout, prior work compresses the ever-growing history in different ways: PackForcing
uses a learned fixed-size memory~\citep{mao2026packforcing}, while FramePack applies recency-weighted,
multi-scale context packing~\citep{zhang2025packing}. These approaches are complementary to our focus on
bidirectional autoregressive training.

\section{Method}
\label{sec:method}

\ours{} converts a pretrained multi-step bidirectional video-diffusion model into a few-step,
camera-controllable, chunk-wise autoregressive world model. We emphasize at the outset that the entire
training pipeline comprises exactly two stages (Fig.~\ref{fig:pipeline}): \textbf{Stage~1}, camera-text
pretraining, and \textbf{Stage~2}, multi-objective few-step distillation. There is no separate data-curation,
or post-alignment stage. The central design choice, illustrated in Fig.~\ref{fig:architecture}, is to
factorize generation into chunks that are denoised with full bidirectional attention internally yet
produced autoregressively, each conditioned on the history of preceding chunks and on a stream of discrete
camera-action tokens. We first establish notation (Sec.~\ref{sec:formulation}) and the data preprocessing that
recovers continuous $6$-DoF poses (Sec.~\ref{sec:data}), then describe camera-text control
(Sec.~\ref{sec:camtext}), the bidirectional autoregressive rollout with sliding-window history conditioning
(Sec.~\ref{sec:bar}), the multi-objective distillation (Sec.~\ref{sec:dmd}), how a single recipe spans
cross-attention, MMDiT, and audio--video backbones (Sec.~\ref{sec:backbones}), and the training budget
(Sec.~\ref{sec:budget}).


\subsection{Problem Formulation}
\label{sec:formulation}
Let $\vx = (\vx^1,\dots,\vx^{T})$ be the latent frames of a video produced by the VAE encoder of a
foundation backbone, and let $c$ be a (static-only) scene caption. We partition the $T$ latent frames into
$B$ contiguous chunks of $K$ frames each, $\vx = (\vc_1,\dots,\vc_B)$ with $\vc_b = \vx^{(b-1)K+1:bK}$.
Autoregression simply means generating these chunks one after another: both the causal and the
bidirectional paradigm share the same chunk-wise factorization
\begin{equation}
p(\vx\mid c,\, \va) \;=\; \prod_{b=1}^{B} p\big(\vc_b \;\big|\; \vc_{<b},\, c,\, \va_b\big),
\label{eq:factorization}
\end{equation}
where $\va = (\va_1,\dots,\va_B)$ is a stream of discrete camera actions and $\va_b$ is the per-frame action
sequence governing chunk $b$. What separates the two paradigms is not the chunk size but how each
factor treats the history $\vc_{<b}$ inside the attention.

\paragraph{Causal vs.\ bidirectional autoregression.} A causal autoregressive model imposes a causal
attention mask: while denoising chunk $\vc_b$, the history is read from a frozen key--value cache, and the
current chunk may attend to the past but the past may never attend to the present. Consequently the
representation (the ``state'') of each history frame is fixed the moment it is produced and can never be revised.
\ours{} is instead bidirectional autoregressive: at each step it attends jointly and
bidirectionally over the current chunk and its history, so the state of the history is itself
conditioned on---and refreshed by---the chunk being generated. Because every already-generated frame remains
free to update under the influence of the frames that follow it, the model continually re-interprets and
self-corrects its own past, which is exactly what suppresses the error accumulation and camera drift of strict
causality (Fig.~\ref{fig:causal}). In short, what makes a model causal or bidirectional is simply whether
its already-generated frames are still allowed to change---not how many frames it produces at a time. \ours{}
generates a short chunk of frames at each step and lets that chunk, together with the visible history, attend
back and forth freely; the history is therefore re-encoded at every step and keeps being refined as generation
moves forward. This is modestly more costly than caching the frozen history, but it is what lets the model stay
sharp and on-trajectory while the rollout continues for arbitrarily long.

\subsection{Data Preprocessing: Camera Trajectories to Discrete Actions}
\label{sec:data}
\ours{} is grounded in continuous camera geometry: the discrete control of Sec.~\ref{sec:camtext} is a
quantization of true $6$-DoF poses, not a hand-assigned categorical label. We draw on two complementary sources,
each providing per-frame continuous camera trajectories that are later quantized into the action vocabulary.

\textbf{Prescribed trajectories (OpenVid\,$+$\,WorldPlay).} We directly reuse the open-source prescribed-trajectory
data released by minWM~\citep{zhao2026minwm}, which samples still images from OpenVid~\citep{nan2025openvid} and
uses WorldPlay~\citep{sun2025worldplay} to generate videos that follow specified camera trajectories. Because
the trajectory is prescribed rather than estimated, these clips carry exact ground-truth $6$-DoF poses by
construction, providing clean and diverse camera supervision at scale.

\textbf{Real footage (Sekai).} For real-world coverage we use the Sekai walking dataset~\citep{li2026sekai} as our
real-footage split: in-the-wild egocentric footage whose camera trajectory is not given and must be
recovered. We follow the camera-annotation pipeline of SANA-WM~\citep{zhu2026sana}, running a SLAM-style video
pose engine~\citep{huang2025vipe} grounded with learned monocular geometry---a temporally consistent multi-view
estimator~\citep{wang2025pi} for structure together with a metric monocular model~\citep{wang2026moge} for
absolute scale---and refining per-frame intrinsics through bundle adjustment. This recovers metric-scale per-frame
camera-to-world extrinsics $T_i^{cw}\!\in\!SE(3)$ and intrinsics $K_i$ for every real clip.

\textbf{Filtering and captioning.} Clips pass generic visual filters (aesthetic quality, motion magnitude,
optical-flow consistency, scene-cut removal) and camera-specific filters on field of view, focal-length
consistency, trajectory smoothness, and scale stability, which discard clips whose camera geometry is unreliable.
Captions are written under a strict static-only instruction that describes objects, layout, and appearance but
never camera motion, so that textual supervision cannot leak the trajectory and all motion is learned through the
action stream.

\textbf{From continuous poses to discrete actions.} Finally, the recovered continuous trajectory is converted
into the per-frame relative pose $(\Delta\mathbf{t}_i,\Delta\mathbf{R}_i)$ used by the quantizer of
Sec.~\ref{sec:camtext} (Eq.~\ref{eq:action}). We stress the relationship: \ours{} remains faithful to continuous
$6$-DoF geometry throughout annotation, and discreteness enters only at the final quantization step, which maps
the continuous motion onto the compact $81$-class vocabulary so that it can be expressed as injectable text. The
discrete vocabulary is thus a low-bandwidth, text-friendly encoding of real camera geometry, not a
replacement for it.

\subsection{Text-based Camera Control}
\label{sec:camtext}
The defining choice of \ours{} is to treat camera control as a pure conditioning task carried entirely in
the text space, adding no camera encoder and no new learnable parameters. Following the discrete camera-action
encoding used in Yume-1.5~\citep{mao2026yume1}, we quantize continuous $6$-DoF camera motion into a
translation$\times$rotation vocabulary. Related systems such as HunyuanWorld-1.5~\citep{hunyuanworld2025hy}
inject the resulting action into the diffusion time embedding. We instead express each action as ordinary text
and inject it through the backbone's existing text-conditioning path, which leaves the pretrained input
distribution intact and makes fine-tuning substantially more stable and data-efficient.
The mechanism has four parts:
quantization of camera motion into a discrete action vocabulary, one-time pre-encoding of that vocabulary,
per-frame assembly of a camera-text\,$+$\,caption condition, and per-frame injection through the backbone's
existing cross-attention. Let $\tau(\cdot)$ denote the (frozen) text encoder with output dimension $d_t$, and
let $\mathbf{W}\!:\mathbb{R}^{d_t}\!\to\!\mathbb{R}^{d}$ be the backbone's existing text-projection (the same one
applied to captions).

\begin{figure}[t]
    \centering
    \includegraphics[width=0.62\columnwidth]{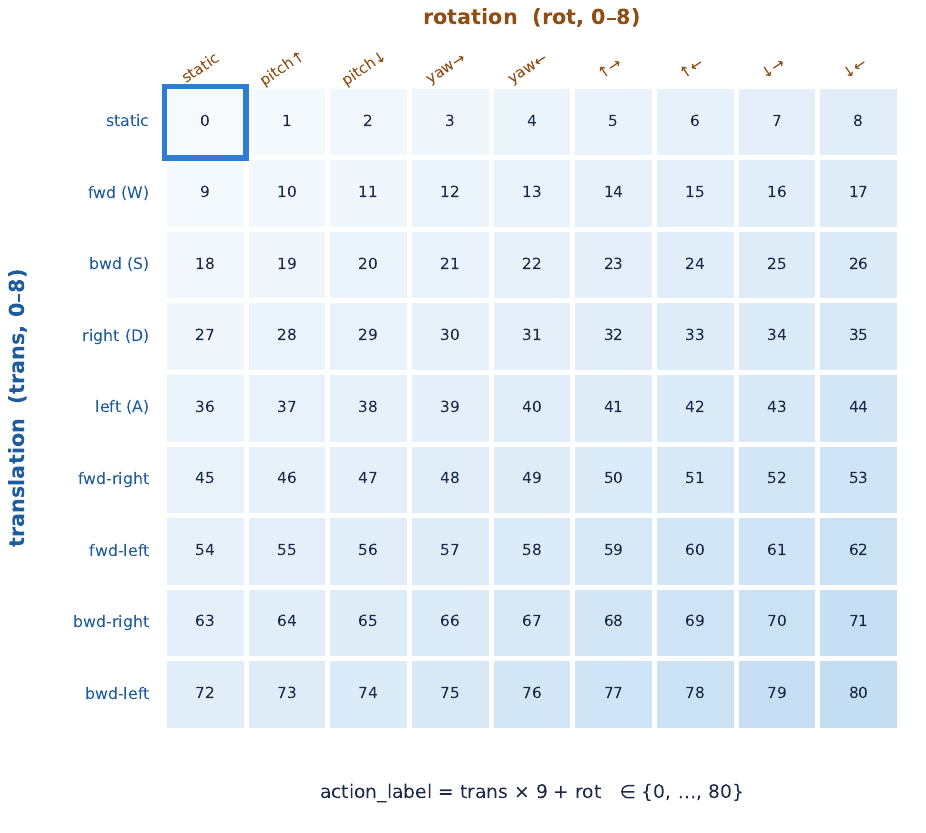}
    \vspace{-4pt}
    \caption{\textbf{The $81$-class discrete camera vocabulary} as a $9\times9$ grid of translation
    $\times$ rotation; each cell is one action label (Eq.~\ref{eq:action}) and maps to a fixed
    camera-text phrase $\phi(a)$.}
    \label{fig:action_grid}
\end{figure}

\textbf{(i) Quantization of camera motion.} We map continuous camera motion to a discrete action per
latent frame. From the continuous trajectory recovered in Sec.~\ref{sec:data}, a frame's relative camera pose,
a translation $\Delta\mathbf{t}_i$ and rotation $\Delta\mathbf{R}_i$ w.r.t.\ the previous keyframe, is fed to a
direction-angle classifier, which assigns a translation class
$g_{\mathrm{t}}(\Delta\mathbf{t}_i)\!\in\!\{0,\dots,8\}$ and a rotation class
$g_{\mathrm{r}}(\Delta\mathbf{R}_i)\!\in\!\{0,\dots,8\}$ (class $0$ = static when the magnitude is below an
adaptive threshold; otherwise the nearest of eight canonical directions), combined into a single label
\begin{equation}
a_i \;=\; \underbrace{g_{\mathrm{t}}(\Delta\mathbf{t}_i)}_{\text{translation}\,\in\,\{0,\dots,8\}}\times 9
        \;+\; \underbrace{g_{\mathrm{r}}(\Delta\mathbf{R}_i)}_{\text{rotation}\,\in\,\{0,\dots,8\}}
        \;\in\; \{0,\dots,80\},
\label{eq:action}
\end{equation}
i.e.\ a $9$-way translation (static / forward / backward / left / right / four diagonals) crossed with a $9$-way
rotation (static / pitch$\,\pm$ / yaw$\,\pm$ / four diagonals); see Fig.~\ref{fig:action_grid}. The
direction--magnitude decoupling makes the quantization robust to pose noise. Equivalently, keyboard/mouse logs or
a textual pose string are parsed directly to the same labels, yielding a per-clip action stream
$\mathbf{a}=(a_1,\dots,a_{T_{\!a}})$.

\textbf{(ii) One-time pre-encoding (separate encoding).} Each of the $81$ classes is tied to a fixed
natural-language camera-motion phrase $\phi(a)$ (e.g.\ ``\textit{Camera moves forward. Camera yaws right.}''). Crucially,
the camera vocabulary and the scene caption are encoded separately, and the vocabulary is encoded
once at initialization rather than every step:
\begin{equation}
\mathbf{E}[a] \;=\; \tau\big(\phi(a)\big)\in\mathbb{R}^{L_a\times d_t},\quad a\in\{0,\dots,80\},
\qquad
\widehat{\mathbf{E}}\in\mathbb{R}^{81\times S_a\times d_t}\ \text{(zero-padded to}\ S_a=\max_a L_a\text{)} ,
\label{eq:preencode}
\end{equation}
and stored as a frozen buffer. At run time we only gather from $\widehat{\mathbf{E}}$; the text encoder is
never invoked on camera phrases during training, which removes them from the per-step training cost.

\textbf{(iii) Per-frame condition assembly (text-feature concatenation).} Let $\mathbf{C}=\tau(c)\in
\mathbb{R}^{S_c\times d_t}$ be the caption embedding. For each latent frame $i$ we concatenate its gathered
camera-text with the caption and project through the shared text head,
\begin{equation}
\mathbf{Z}_i \;=\; \Pi_{S}\!\big(\big[\,\widehat{\mathbf{E}}[a_i]\ ;\ \mathbf{C}\,\big]\big)\in\mathbb{R}^{S\times d_t},
\qquad
\mathbf{H}_i \;=\; \mathbf{W}\,\mathbf{Z}_i \in\mathbb{R}^{S\times d},
\label{eq:assemble}
\end{equation}
where $[\,\cdot\,;\,\cdot\,]$ is row-wise concatenation and $\Pi_S$ truncates/zero-pads to the fixed text length
$S$ (we use $S{=}512$). Because the cross-attention applies no positional encoding to the context and
no key masking, the result is invariant to the order of the camera and caption rows, which permits
implementing Eq.~\ref{eq:assemble} as a single padded, vectorized gather (camera-to-latent length is aligned by
nearest-neighbour interpolation, $\mathbf{a}\!\leftarrow\!\mathrm{NN}(\mathbf{a},T)$, which preserves discrete
class boundaries that linear interpolation would blur).

\textbf{(iv) Per-frame injection (decoupled streams).} At every block, the camera-text enters through
the backbone's existing cross-attention, with no new module. Each latent frame being generated is reshaped
per frame and attends to its own condition: denoting frame $i$'s $N_p$ patch tokens by $\mathbf{X}_i$,
\begin{equation}
\mathbf{X}_i \;\mathrel{+}=\; \mathrm{CrossAttn}\big(\,\underbrace{\mathbf{X}_i}_{\text{query}},\ \underbrace{\mathbf{H}_i}_{\text{key/value}}\,\big),
\label{eq:inject}
\end{equation}
so each frame attends to its own camera-text\,$+$\,caption $\mathbf{H}_i$. In Stage~1 the whole clip is denoised
jointly and there is no history, so this per-frame injection is the only conditioning path: every latent frame
receives its prescribed camera action through Eq.~\ref{eq:inject}.

During autoregressive rollout (Stage~2, Sec.~\ref{sec:bar}), the already-generated frames are summarized into
history/memory tokens $\mathbf{X}^{\mathrm{mem}}$ (produced by one of the three history modes below) that condition the next chunk. These memory tokens carry
no camera-text and attend to the caption only,
\begin{equation}
\mathbf{X}^{\mathrm{mem}} \;\mathrel{+}=\; \mathrm{CrossAttn}\big(\mathbf{X}^{\mathrm{mem}},\ \mathbf{W}\,[\,\mathbf{C};\mathbf{0}\,]\big),
\label{eq:inject-mem}
\end{equation}
so the camera action is applied only to the chunk currently being generated, never re-applied to history.
Keeping the two streams separate disentangles ``what the world looks like'' (caption) from ``how the camera
moves'' (action), and prevents the history from leaking spurious camera cues into the future.

\textbf{Parameter efficiency and training stability.} Every operation above reuses pretrained components: the gather from
$\widehat{\mathbf{E}}$ is parameter-free, and $\mathbf{W}$ and the cross-attention are the backbone's own. Unlike
absolute or residual-relative injection, \ours{} adds no parameters and no residual branch onto the
self-attention, so at step $0$ the model is exactly the pretrained generator conditioned on richer text. This is
why control emerges in $\sim\!100$ steps (Sec.~\ref{sec:budget}) without the early-training quality dip or the
large-batch requirement that prior residual camera-injection methods rely on.

\subsection{Chunk-wise Autoregressive Rollout and Sliding-Window History Conditioning}
\label{sec:bar}
To realize Eq.~\ref{eq:factorization}, chunk $b$ is generated by the backbone conditioned on a memory
representation of the already-generated history $\vc_{<b}$. The already-generated clean latents are
placed at noise level $\sigma{=}0$ in the sequence prefix, and the new chunk denoises conditioned on them through
the backbone's native image-to-video timestep separation. To bound the cost, conditioning is restricted to a
sliding window of the most recent clean latents together with a first-frame sink that anchors global layout;
history beyond the window is discarded rather than compressed. The scheme is exact within the window and
parameter-free, which makes it suitable for short to medium rollouts. The prefix retains the backbone's native
patch-grid coordinates and true temporal indices for RoPE, avoiding spatial jumps or temporal-order changes at
chunk boundaries. Because no compressed long-range memory is retained, distant context is lost once the rollout
outgrows the window.

\subsection{Multi-Objective Few-Step Distillation}
\label{sec:dmd}
The camera-text-pretrained model of Sec.~\ref{sec:camtext} is a $50$-step bidirectional sampler. Stage~2 distills
it into a $4$-step chunk-wise generator with self-rollout distribution-matching distillation. The training
loop directly builds on Self-Forcing~\citep{huang2026self}---the generator rolls out its own chunks during
training and is supervised by a DMD objective, closing the train--test gap of autoregressive video diffusion---and
our key change is the paradigm: where Self-Forcing rolls out under a strictly causal mask, \ours{}
rolls out chunk-wise bidirectionally (full attention within each chunk, autoregression across chunks,
Eq.~\ref{eq:factorization}). We use three copies initialized from the Stage-1 weights: a frozen
real score $s_\text{real}$ (the bidirectional teacher, evaluated with classifier-free guidance), an online
fake score $s_\text{fake}$ (a critic tracking the student's distribution), and the generator
$G_\theta$ with velocity field $v_\theta$, which self-rolls out a sequence $\tilde\vx$ chunk by chunk. We adopt a flow-matching parameterization: for a clean latent $\vx_0$, noise level
$\sigma\!\in\!(0,1)$ and $\vepsilon\!\sim\!\mathcal{N}(0,I)$, the noised sample is
$\vx_\sigma=(1-\sigma)\vx_0+\sigma\vepsilon$, the velocity target is $(\vepsilon-\vx_0)$, and the model's clean
estimate is $\hat\vx_0=\vx_\sigma-\sigma\,v_\theta(\vx_\sigma,\sigma,c,\va)$. The Stage-2 objective is a primary
distribution-matching term regularized by two families of complementary anchors.

\textbf{Primary: distribution-matching distillation.} The leading term aligns the student to the teacher
distribution through the asymmetric DMD gradient~\citep{yin2024one,yin2025slow,huang2026self}
\begin{equation}
\nabla_\theta\, \mathbb{E}_t\!\big[\KL\!\big(p_{\theta,t}(\tilde\vx_t)\,\|\,p_{\text{data},t}(\tilde\vx_t)\big)\big]
= -\,\mathbb{E}_{\tilde\vx,\,t,\,\tilde\vx_t}\!\Big[\big(s_\text{real}(\tilde\vx_t,t)-s_\text{fake}(\tilde\vx_t,t)\big)\,
\tfrac{\partial \tilde\vx}{\partial \theta}\Big],
\label{eq:dmd}
\end{equation}
where $\tilde\vx_t$ is the noised student sample at level $t$, and $s_\text{real},s_\text{fake}$ receive the same
caption and camera-text conditions so controllability survives distillation. The critic $s_\text{fake}$ is
itself trained online (at an $N{:}1$ ratio against the generator) with a flow-matching velocity loss on the
detached generator outputs. To keep self-rollout tractable, we retain the gradient on a single randomly chosen
denoising step per chunk~\citep{huang2026self} and detach history across chunks, so the graph never spans the
full rollout; a dynamic chunk-count curriculum concentrates compute on longer histories. Crucially,
Eq.~\ref{eq:dmd} is a reverse-KL objective and is therefore mode-seeking: minimized in isolation it
tends to drop modes, manifesting as motion that decays toward a static scene or as high-frequency collapse. The two
anchor families below counteract these failure modes.

\textbf{Supervised anchor (SFT, low-$\sigma$ velocity MLE).} On the full real video latent $\vx_0$
(all $T$ frames, decoupled from the per-iteration rollout length) we add a flow-matching velocity regression at low
noise levels:
\begin{equation}
\mathcal{L}_{\text{SFT}}(\theta)=\mathbb{E}_{\vx_0,\,\sigma\sim\mathcal{U}(\sigma_{\min},\,\sigma_{\text{sft}}),\,\vepsilon}
\big[\,\big\|\,v_\theta(\vx_\sigma,\sigma,c,\va)-(\vepsilon-\vx_0)\,\big\|^2\,\big],
\qquad \sigma_{\text{sft}}\ \text{small}.
\label{eq:sft}
\end{equation}
This design is inspired by the regression regularizer in the original DMD formulation~\citep{yin2024one},
which aligns the distilled model with teacher samples to mitigate the mode-seeking behavior of distribution
matching. We instead align the student directly with real videos. At low $\sigma$, this provides a strong
maximum-likelihood anchor that both reduces mode dropping and improves visual fidelity. It also allows
practitioners to use a curated set of high-quality videos as supervised anchors during DMD, steering the
distilled model toward the desired data distribution. Because the loss is computed on the complete video rather
than on the (possibly short) rollout, it preserves long-video and motion modeling even when warmup uses a single
block.

\textbf{Forward-KL anchors (mass-covering).} Our construction is inspired by the teacher-sample alignment in
the original DMD formulation~\citep{yin2024one} and related findings in HiAR~\citep{zou2026hiar}. Both motivate
anchoring the student to samples from a coverage-oriented reference distribution rather than relying solely on
the mode-seeking reverse-KL DMD gradient. Building on these ideas, we adapt the objective to bidirectional
autoregressive video and make task-specific refinements to the anchor construction, noise ranges, and loss
weights. We therefore add a forward-KL term
$\KL(p_{\text{data}}\,\|\,p_\theta)$, which is mass-covering and penalizes dropping data modes (the source of
low-motion degeneration). Minimizing forward KL reduces to an $\vx_0$-regression (maximum-likelihood) objective
on samples from the covered distribution, which we instantiate two ways.
(a) Real forward-KL noises the full real video at high $\sigma$ and regresses the student's clean
estimate back to it,
\begin{equation}
\mathcal{L}_{\text{rFKL}}(\theta)=\mathbb{E}_{\vx_0,\,\sigma\sim\mathcal{U}(\sigma_{\text{lo}},\,\sigma_{\text{hi}}),\,\vepsilon}
\big[\,\big\|\,\hat\vx_0(\vx_\sigma,\sigma)-\vx_0\,\big\|^2\,\big],
\qquad \sigma_{\text{lo}}\!>\!\sigma_{\text{sft}},
\label{eq:rfkl}
\end{equation}
complementing SFT: high $\sigma$ governs global layout and motion, low $\sigma$ governs detail. (b) Teacher
forward-KL is data-free: the frozen teacher (with CFG) is rolled out along a dense ODE trajectory
$\sigma{:}1\!\to\!0$; for sampled trajectory anchors $(\vx_{\sigma_a},\vx_{\sigma_b})$ we read off the teacher's
clean target by linear extrapolation, $\vx_0^{\text{teach}}=\vx_{\sigma_a}-\sigma_a(\vx_{\sigma_b}-\vx_{\sigma_a})/(\sigma_b-\sigma_a)$,
and regress the student's estimate at the same point:
\begin{equation}
\mathcal{L}_{\text{tFKL}}(\theta)=\mathbb{E}\big[\,\big\|\,\big(\vx_{\sigma_a}-\sigma_a\,v_\theta(\vx_{\sigma_a},\sigma_a,c,\va)\big)-\vx_0^{\text{teach}}\,\big\|^2\,\big].
\label{eq:tfkl}
\end{equation}
This transfers the teacher's full, mass-covering distribution without any real data, countering mode shrink and
preserving rich camera-driven motion.

\textbf{Total objective.} The generator minimizes
\begin{equation}
\mathcal{L}=\mathcal{L}_{\text{DMD}}
+\lambda_{\text{SFT}}\mathcal{L}_{\text{SFT}}
+\lambda_{\text{rFKL}}\mathcal{L}_{\text{rFKL}}
+\lambda_{\text{tFKL}}\mathcal{L}_{\text{tFKL}},
\label{eq:total}
\end{equation}
where each auxiliary term is optional and toggled by a single flag, letting practitioners trade stability for
speed. Conceptually, DMD matches the teacher, the forward-KL anchors restore coverage, and SFT anchors fine detail
to real data. With this objective the generator produces each $K$-frame chunk in $4$ denoising steps
and rolls out to $60$\,s and beyond.

\begin{figure*}[!t]
    \centering
    \includegraphics[width=\textwidth]{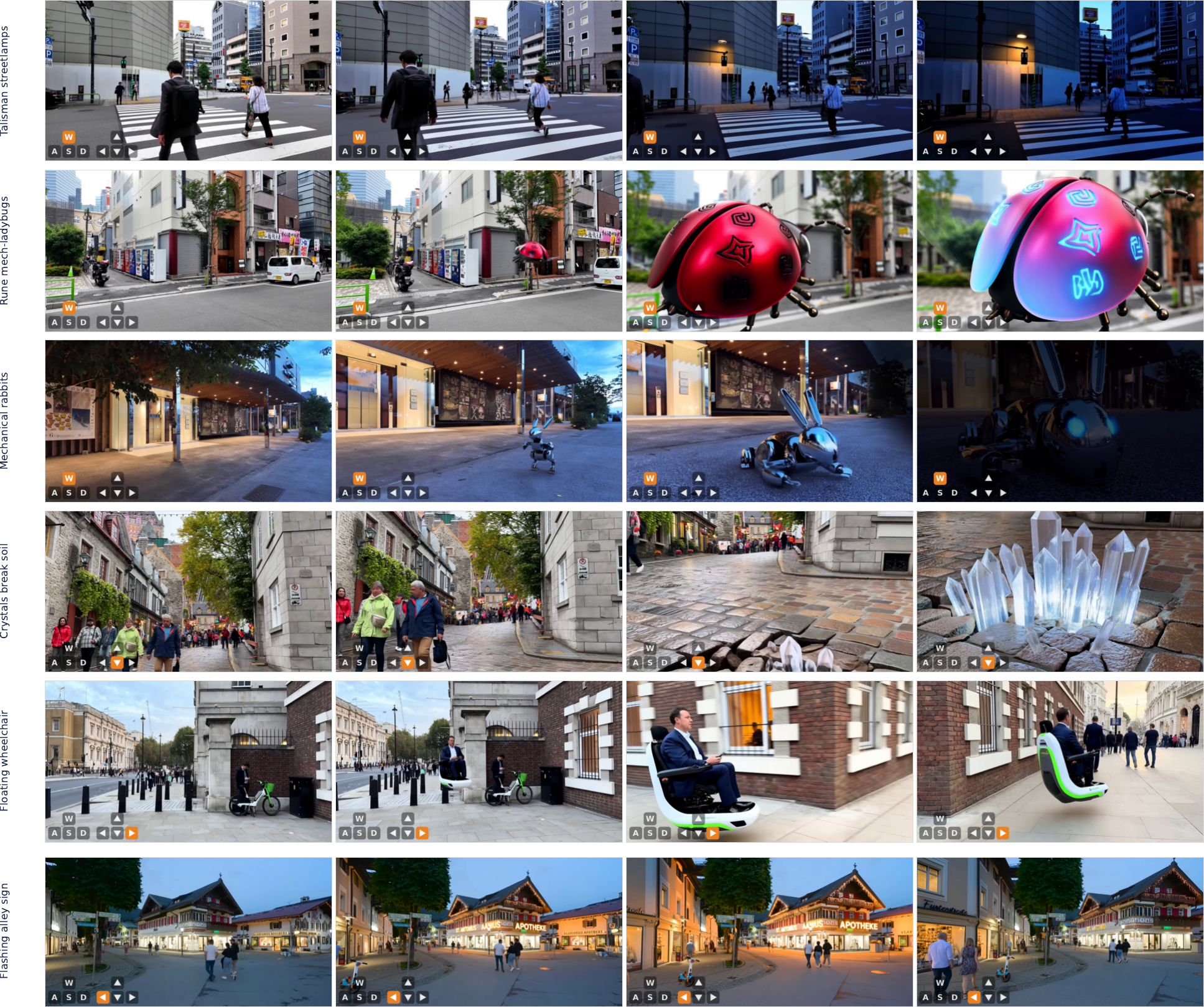}
    \vspace{-6pt}
    \caption{\textbf{Event generation / real-time event editing.} \ours{} injects prompt-specified, fantastical
    events into real street scenes while the camera moves (joystick overlay, bottom-left). Top to bottom: glowing
    talisman streetlamps, rune-covered mechanical ladybugs that repel insects, mechanical rabbits, crystal clusters
    breaking through the soil, a self-driving floating wheelchair, and a flashing alley advertisement sign; each is
    shown over four frames. The event is specified purely by text and can be introduced or switched mid-rollout in
    real time. We release the event dataset and scripts. Qualitative illustration, not a benchmark.}
    \label{fig:events}
\end{figure*}

\textbf{Real-time event editing.} A capability that the bidirectional paradigm makes natural---and that, to our
knowledge, no prior open framework releases---is \emph{event editing}: injecting a textual \emph{event} into the
scene while it is being explored. Each chunk is conditioned jointly on the event text and the discrete camera
action, and because the history is continually re-encoded (Sec.~\ref{sec:formulation}), a user can introduce an
event for the upcoming chunks and the bidirectional self-correction weaves it into the ongoing world coherently
and in real time, then move on to the next event seamlessly. Fig.~\ref{fig:events} shows \ours{} realizing
fantastical, prompt-specified events---glowing talisman streetlamps, rune-covered mechanical ladybugs, crystals
breaking through the soil, a self-driving floating wheelchair---inside real street scenes while the camera moves
under the joystick overlay. \ours{} exposes this as a first-class feature, and we release the event dataset and
scripts alongside the framework.

\textbf{Image-to-video capability.} Image-to-video capability is not obtained for free from a
text-to-video checkpoint. For a T2V backbone, Stage~1 must include image-conditioned training so that the model
learns to use the input frame together with the camera-action stream. At inference, the user-provided image is
encoded into the first clean latent frame and placed as the initial history, after which the model continues the
sequence under camera control. Native TI2V backbones such as Wan2.2-TI2V-5B already provide an image-conditioning
path; for these models, Stage~1 retains and adapts the pretrained capability rather than learning it from
scratch, and Stage~2 preserves it during distillation. The framework also supports mixed-task Stage~1 training over
text-to-video, image-to-video, and video-to-video data when a shared checkpoint across these modes is desired.

\subsection{Generality Across Architectures and Modalities}
\label{sec:backbones}
Nothing above is tied to a particular backbone: \ours{} only requires a video-diffusion model whose attention can
be evaluated chunk-wise and whose conditioning accepts per-frame text. We exploit this to instantiate the same
two-stage recipe across three architecture families. On cross-attention backbones (Wan2.1-T2V-1.3B and
Wan2.2-TI2V-5B~\citep{wan2025wan}), camera-text enters through the existing text cross-attention. On an
MMDiT backbone (HunyuanVideo-1.5~\citep{kong2024hunyuanvideo,esser2024scaling}), where text and video
tokens are jointly attended in double-stream blocks, the camera-text tokens are concatenated into the text stream
and the chunk/history logic wraps the joint attention. On a joint audio--video backbone
(LTX-2.3-22B~\citep{hacohen2024ltx}), the chunk groups paired audio and video latents so that each window denoises
synchronized sound and vision together; the across-chunk history carries both streams, yielding an interactive
world model that is audible as well as visible. Adapting to a new backbone amounts to providing an encoder
for clean-latent history and a hook for per-frame camera-text, typically a thin adapter, while the camera
vocabulary, rollout, and distillation code are shared.

\subsection{Training Budget}
\label{sec:budget}
A practical highlight of \ours{} is how little training it needs. The chunk-wise bidirectional design keeps the
backbone close to its pretrained prior, and the disentangled discrete camera-text is a low-dimensional signal to
learn, so Stage~1 acquires reliable camera control in only $\sim\!100$ optimizer steps. Stage~2 distillation,
with the SFT anchor of Eq.~\ref{eq:sft}, converges in $\sim\!200$ steps. Both stages run on $8\times$H200 GPUs
with gradient accumulation $4$, so the entire pipeline completes in hours rather than days. Notably, there is no separate
post-alignment stage; the two stages above constitute the whole recipe.

\section{Implementation Details}
\label{sec:implementation}

We summarize the settings needed to reproduce \ours{} beyond the recipe of Sec.~\ref{sec:method}. Real clips are
truncated to $77$ frames and captioned under a static-only instruction so that motion is carried solely by the
discrete action stream; captions and the per-class camera-text bank are pre-encoded once for efficiency, and
action-to-latent length alignment uses nearest-neighbour interpolation to preserve discrete class boundaries. In
Stage~2 the generator self-rolls out chunk by chunk with one random denoising step per chunk retaining gradient
(history detached across chunks), and a dynamic chunk-count curriculum concentrates compute on longer histories.
Both stages run on $8\times$H200 GPUs with gradient accumulation $4$, converging in $\sim\!100$ (Stage~1) and
$\sim\!200$ (Stage~2) optimizer steps; each auxiliary loss is toggled by a single flag. For full reproducibility,
all qualitative figures in this paper are generated from raw rollout
frames by the released figure-generation script, and the released fine-tuning scripts reproduce both training
stages and inference across the four backbones.

\section{Results}
\label{sec:components}
Because \ours{} is a framework rather than a single model, we characterize it through the function of each
design choice rather than through a single benchmark number. We first record how the framework is instantiated and
then analyze, component by component, the contribution and rationale of each element. A systematic quantitative
study across backbones is ongoing and will accompany the code release; the goal here is to make the role of every
component precise.

\subsection{Instantiations}
\label{sec:instantiations}
\textbf{Backbones.} The same recipe is instantiated on four backbones spanning three architecture families:
cross-attention condition injection (Wan2.1-T2V-1.3B and Wan2.2-TI2V-5B~\citep{wan2025wan}), an MMDiT design
(HunyuanVideo-1.5~\citep{kong2024hunyuanvideo}), and a joint audio--video design (LTX-2.3-22B~\citep{hacohen2024ltx}).
Unless noted, clips are $77$ frames (encoded to $20$ latent frames) at each backbone's native resolution, grouped
into chunks of $K$ latent frames, with the distilled generator run at $4$ denoising steps per chunk.

\textbf{Data.} Both stages share one format, a per-clip caption plus discrete camera actions: OpenVid$+$WorldPlay
clips with prescribed-trajectory actions (Sec.~\ref{sec:data}), and the real-footage (Sekai) split,
whose recovered poses are quantized into the $81$ combined-action classes ($9$ translation $\times$ $9$ rotation).
Captions are produced by a vision-language
model under a strict static-only instruction that describes scene appearance but never camera or object motion, so all
motion supervision flows through the discrete action stream. These real pairs also supply the SFT and real
forward-KL targets (Eqs.~\ref{eq:sft},~\ref{eq:rfkl}).

\textbf{Training budget.} Both stages are short---$\sim\!100$ (Stage~1) and $\sim\!200$ (Stage~2) optimizer
steps on $8\times$H200 GPUs with gradient accumulation $4$, with no separate alignment stage; see
Sec.~\ref{sec:budget} for why this short budget suffices.

\subsection{Qualitative Results}
\label{sec:roles}

These results illustrate \ours{}'s qualitative behavior; they convey mechanism and visual quality rather than
serving as quantitative benchmarks.

\begin{figure*}[!t]
    \centering
    \includegraphics[width=\textwidth]{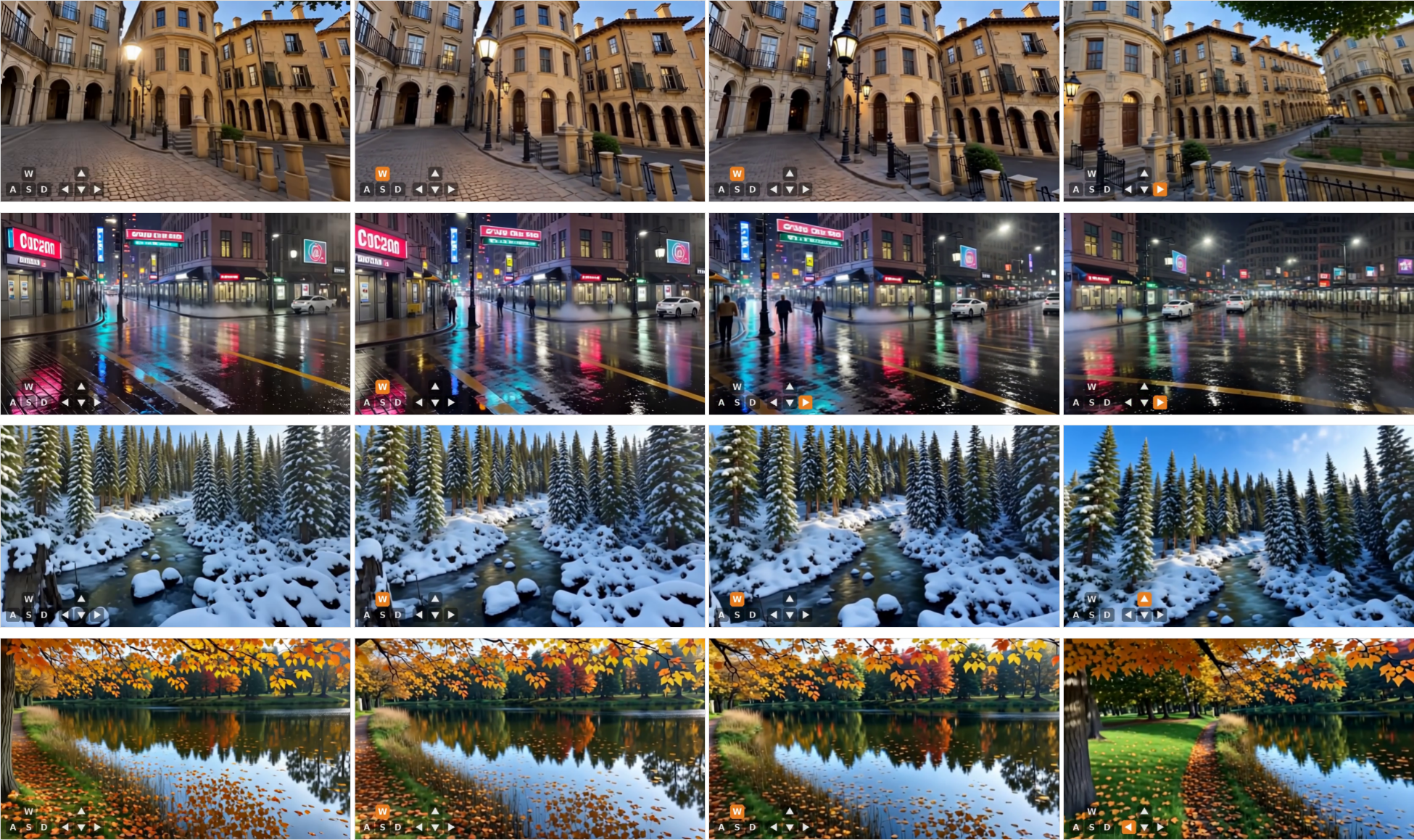}
    \vspace{-6pt}
    \caption{\textbf{Illustrative T2V rollouts under camera control.} Each row is a different text-prompted scene
    rolled out chunk-wise under its own camera motion (joystick overlay, bottom-left); keeping each window
    bidirectional preserves scene identity and geometry as the camera moves. Shown to illustrate the mechanism,
    not as a quantitative evaluation.}
    \label{fig:t2v}
\end{figure*}

\textbf{Long-horizon rollouts under camera control.} Figure~\ref{fig:t2v} shows chunk-wise text-to-video rollouts
in which each window stays bidirectional and the history keeps updating as it is generated
(Sec.~\ref{sec:formulation},~\ref{sec:bar}). Scene identity and geometry are preserved as the camera moves, and
the visible history is re-encoded at every chunk.

\textbf{Camera controllability.} Figure~\ref{fig:camera_control} isolates the text-based control of
Sec.~\ref{sec:camtext}: with a single constant discrete action per row, the camera obeys the prescribed
translation and look direction while preserving scene fidelity.

\begin{figure*}[!t]
    \centering
    \includegraphics[width=\textwidth]{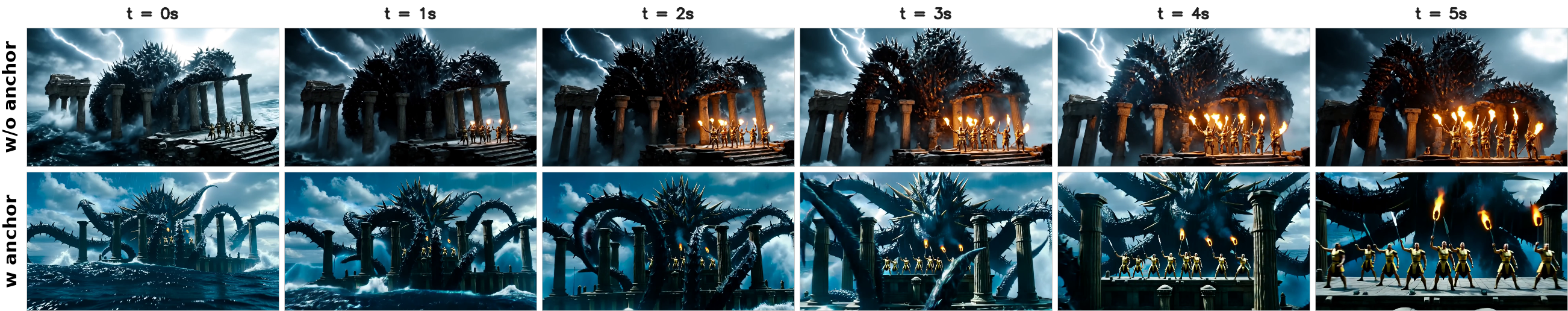}
    \vspace{-6pt}
    \caption{\textbf{With vs.\ without the anchor losses.} Same prompt, camera trajectory, and
    random seed; frames sampled every second from a $5$\,s rollout. The top row (w/o anchor loss) is the
    $4$-step generator distilled with the DMD term only; the bottom row (w/ anchor loss) adds the SFT and
    forward-KL anchors. The anchor losses yield markedly sharper detail, higher contrast, and richer temporal
    dynamics, whereas DMD alone drifts toward a hazy, near-static rollout.}
    \label{fig:ablation_anchors}
\end{figure*}

\textbf{Effect of the anchor losses.} Figure~\ref{fig:ablation_anchors} contrasts the generator distilled
without the anchor losses (DMD term alone) against the one trained with them (adding the SFT and forward-KL
anchors), under the same prompt, camera script, and random seed. Without the anchor losses the
rollout is hazy and low in contrast, and its content barely changes over time, a direct symptom of the
mode-seeking bias toward static, over-smoothed motion. With the anchor losses, high-frequency structure and
contrast are restored and temporal dynamics increase markedly, with lighting and scene geometry evolving visibly
across the horizon. The terms are complementary: the SFT anchor ties fine detail to real data, while the
forward-KL anchors preserve motion coverage.

\section{Conclusion}
We presented \ours{}, a recipe for bidirectional autoregressive video world models built on a
chunk-wise factorization that retains the backbone's full bidirectional attention within each generated chunk
while rolling out autoregressively across chunks. Two short training stages, camera-text pretraining and a
multi-objective few-step distillation that augments distribution matching with auxiliary SFT and forward-KL
terms, transform a $50$-step bidirectional teacher into a $4$-step chunk-wise generator steered by an $81$-class
discrete camera-action vocabulary. The recipe is economical: control emerges in $\sim\!100$ steps and
distillation converges in $\sim\!200$ steps on $8\times$H200 GPUs, with no separate alignment stage. It is
also broad: a single recipe spans cross-attention (Wan2.1-1.3B, Wan2.2-5B), MMDiT (HunyuanVideo-1.5), and
audio--video (LTX-2.3-22B) backbones, with the last yielding a world model that generates synchronized audio together with
vision. For T2V backbones, image-to-video capability requires image-conditioned Stage~1 training, whereas native
TI2V backbones such as Wan2.2-5B retain their pretrained image-conditioning path. We
regard \ours{} as the bidirectional point in the same design space as causal frameworks such as minWM, trading a
small amount of per-window latency for fidelity, controllability, and markedly shorter training. Future
directions include continuous and compositional action vocabularies, stronger long-horizon memory, richer
audio--video control, and head-to-head benchmarking against causal recipes; we release code, checkpoints, and
inference scripts to support them.

\clearpage
\newpage
\bibliographystyle{assets/plainnat}
\bibliography{ref}

\end{document}